\documentclass[conference]{ieeeconf}      

\IEEEoverridecommandlockouts                              

\usepackage{graphicx}
\usepackage{subfig}
\usepackage{amsmath} 
\usepackage{amssymb}  
\usepackage{bm} 
\usepackage{mathtools} 

\usepackage[ruled,norelsize]{algorithm2e}

\makeatletter
\newcommand{\removelatexerror}{\let\@latex@error\@gobble}
\makeatother

\title{\LARGE \bf
Learning to Interrupt: A Hierarchical Deep Reinforcement Learning Framework for Efficient Exploration
}

\author{Tingguang Li, Jin Pan, Delong Zhu, Max Q.-H. Meng* \\
	The Department of Electronic Engineering, 
	The Chinese University of Hong Kong, \\
	Shatin, N.T., Hong Kong SAR, China \\
    Email: \{tgli, jpan, dlzhu, qhmeng\}@ee.cuhk.edu.hk	
}

\begin{document}

\maketitle
\thispagestyle{empty}
\pagestyle{empty}
{\let\thefootnote\relax\footnote{{*The corresponding author.}}}
{\let\thefootnote\relax\footnote{{This work is supported by Hong Kong RGC GRF grants \#14200618.}}}

\begin{abstract}
To achieve scenario intelligence, humans must transfer knowledge to robots by developing goal-oriented algorithms, which are sometimes insensitive to dynamically changing environments. While deep reinforcement learning achieves significant success recently, it is still extremely difficult to be deployed in real robots directly. In this paper, we propose a hybrid structure named Option-Interruption in which human knowledge is embedded into a hierarchical reinforcement learning framework. Our architecture has two key components: \emph{options}, represented by existing human-designed methods, can significantly speed up the training process and \emph{interruption mechanism}, based on learnable termination functions, enables our system to quickly respond to the external environment. To implement this architecture, we derive a set of update rules based on policy gradient methods and present a complete training process. In the experiment part, our method is evaluated in Four-room navigation and exploration task, which shows the efficiency and flexibility of our framework.
\end{abstract}

\section{INTRODUCTION}
 For robots at the current stage, it is considered more practical for them to work in a specific environment or solve a specific problem. Such a concept is often referred to as scenario intelligence in contrast to the general artificial intelligence. To achieve such scenario intelligence, humans must transfer knowledge to robots by developing goal-oriented algorithms. Though feasible, these human-designed methods sometimes are insensitive to the dynamically changing environments. 

 Recently, Deep Reinforcement Learning (DRL) has achieved significant success in various games \cite{Mnih2015Human}\cite{Silver2017Mastering} and shows a promising future. However, directly applying such methods to real robots is quite difficult. First of all, the majority of these work are trained via thousands of trial-and-error episodes while the robot is too fragile to sustain such a process. Besides, considering there exist a bunch of algorithms off the shelf in the robotics area, sometimes it's inefficient to make the robot learn from scratch. In this paper, the relationship between the existing human-designed methods and DRL is re-examined. The main concern here is whether we can develop an algorithm that inherits the efficiency and flexibility of learning-based methods and holds a controllable training process on the basis of existing methods. 

\begin{figure}
	\centering
	\includegraphics[height=130pt]{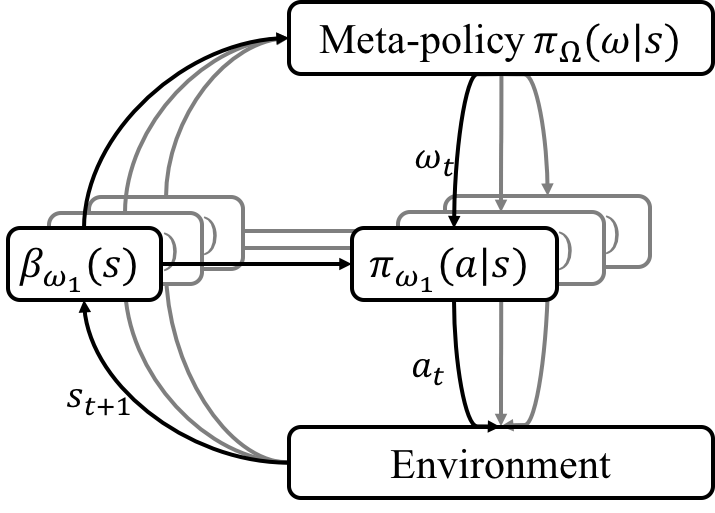}
	\caption{Schematic diagram of Hierarchical Reinforcement Learning (HRL) with three options and option 1 is selected.} \label{fig:HRL}
\end{figure}

Human decision making routinely involves choices over a broad range of time scales. By contrast, the Markov Decision Process (MDP) based Reinforcement Learning (RL) does not have such foresight and makes the action only based on current observation. Facing such limitations, Sutton \cite{sutton1999between} proposed \emph{options} to represent the courses of actions that take place at different time scales. Different from typical RL, this Hierarchical Reinforcement Learning (HRL) architecture is based on Semi-Markov Decision Process (SMDP) and intended to model temporally extended action sequence. 

A schematic diagram of HRL is depicted in Fig. \ref{fig:HRL}. Instead of picking the action directly, the agent first selects an option $\omega$ according to the meta-policy $\pi_\Omega$, then follows its intra-option policy $\pi_\omega$ until being ended by its termination function $\beta_\omega$. Based on this HRL structure, the Option-Interruption framework is proposed in this paper. Inspired by the fact that the intra-option policies are independent to the meta-policy, we encode the options with the existing human-designed methods, augmented with learnable termination functions. Such a combination brings several benefits: 1) By imparting human knowledge, the training process can be significantly sped up; 2) When combined with well-developed algorithms, the robot's behavior can be restrained, for instance, equipped with the obstacle avoidance algorithms, the agent will not hit the wall when learning to navigate; 3) The interruption mechanism enables the system to constantly monitor and make a response to their external environments; 4) Our architecture is flexible and can be embedded in various scenarios with the replacement of options. 

In this paper, we propose the Option-Interruption architecture combining traditional methods with hierarchical reinforcement learning to achieve the scenario intelligence. The update rule of the meta-policy $\pi_\Omega$ is derived and the training process is given. The experimental results verify the effectiveness of our method.

\section{PRELIMINARIES} \label{sec:related}
A finite discounted Markov Decision Process can be denoted as $M\small{=}\{S,A,P,R, \gamma\}$, where $S$ and $A$ represent the set of states and actions respectively. The state transition probability function $P:S\times S \times A\to [0,1]$ is a conditional distribution over next states given that an action $a\in A$ is taken under current state $s\in S$. The $R:S\times A \to \mathbb{R}$ is a reward function and $\gamma\in [0,1]$ is a discount factor. The policy $\pi: S\times A \to [0,1]$ is a probability distribution over actions conditioned on states. The objective of the agent is to learn a policy that maximizes the  state-value function, i.e. the expected discounted future reward: $V_\pi (s)=\mathbb{E}_\pi[\sum_{t=0}^{\infty}\gamma^tr_{t}|s_0=s]$ and similarly its action-value function is defined as $Q_\pi(s,a)=\mathbb{E}_\pi[\sum_{t=0}^{\infty}\gamma^tr_{t}|s_0=s, a_0=a]$. 

MDP is conventionally conceived that it does not involve temporally extended actions \cite{sutton1999between} and hence is unable to take advantage of high-level temporal abstraction. To correct such deficiency, Semi-Markov Decision Process is proposed \cite{sutton1999between}\cite{barto2003recent} for continuous-time discrete-event systems. The actions in SMDP last multiple time steps and are intended to model temporally extended courses of action. Therefore the transition probability generalizes to a joint probability $P(s_{t+1},\tau|s_t,a_t)$, the transition from state $s_t$ to state $s_{t+1}$ occurs after a positive waiting time, denoted by $\tau$, when action $a_t$ is executed.

The idea of temporally extended actions and the hierarchical structure is formulated as the options in \cite{sutton1999between}. A Markovian option $\omega\in\Omega$ can be represented by a triple $(I_\omega, \pi_\omega, \beta_\omega)$, in which $I_\omega \subseteq S$ is an initiation set, $\pi_\omega: S\times A \to [0,1]$ is an intra-option policy, and $\beta_\omega:S\to [0,1]$ is a termination function. The option $\omega$ is available in $s_t$ if and only if $s_t\in I_\omega$. We consider \emph{call-and-return} option execution model \cite{bacon2017option}, in which an agent picks an option $\omega$ according to its meta-policy $\pi_\Omega$, then follows the intra-option policy $\pi_\omega$ until the option terminates stochastically according to $\beta_\omega$ or the option is finished. Thus the value function can be reformulated in $V_\Omega (s) = \sum_{\omega} \pi_\omega (\omega|s) Q_\Omega (s,w)$, where $Q_\Omega(s,w)$ is the option-value function.

Combined with deep learning techniques, such hierarchical architecture has attracted a lot of attention recently. Kulkarni \emph{et al.} \cite{kulkarni2016hierarchical} propose a hierarchical-DQN framework to operate at different temporal scales. By specifying subgoals, a top-level meta-controller learns a policy over intrinsic goals while a lower-level function learns a policy over primitive actions to satisfy subgoals. Bacon \emph{et al.} \cite{bacon2017option} derive a set of policy gradient theorems for options and proposed an option-critic architecture capable of learning both the internal policies and the termination functions, in tandem with the meta-policy, without providing additional rewards or subgoals. Tessler \emph{et al.} \cite{tessler2017deep} present a deep hierarchical approach for lifelong learning and evaluate it in Minecraft game with impressive performance and show a strong ability to reuse knowledge. However, all of these work ignore the fact that there exist a variety of well-developed algorithms off the shelf in robotics field. Learning from scratch is inadvisable for robots since it may cause serious damage to fragile robots.

\section{METHODOLOGY}
In this section, we first introduce the framework of our method, then derive the update rule based on policy gradient methods \cite{sutton2000policy}. 

\subsection{Option-Interruption Framework}
As introduced in Section \ref{sec:related}, a complete decision-making process in HRL involves two stages: select an option $\omega$ according to meta-policy $\pi_\Omega$, then follow the chosen option's policy $\pi_\omega$ until termination. Each option $w$ is composed of $(I_\omega, \pi_\omega, \beta_\omega)$, representing the initialization set, intra-option policy and termination function, respectively. In our Option-Interruption framework, the intra-option policy $\pi_\omega$ is embedded with existing methods and thus deterministic, representing the temporal abstract of the human knowledge. The meta-policy $\pi_{\Omega}$ and termination functions $\beta_{\omega}$ are obtained by training. The intra-option policy $\pi_\omega$ can be flexibly replaced by various existing methods on the needs of the scenario. 

Take autonomous exploration as an example, where the robot is expected to explore the whole environment with a path as short as possible. Strategies of training the agent with primitive actions, e.g. \emph{up, down, left} and \emph{right} are quite inefficient, not to mention the frequent collision with obstacles which further increases the training difficulties. By substituting options with path planning algorithms like $A^*$ algorithm, the agent can focus on high-level decision making and ignore low-level collison checking, thus making the training process more efficient.

\subsection{Update Rules}
In this subsection, we derive how to train our framework utilizing policy gradient methods. Policy gradient methods \cite{sutton2000policy} learn the policy parameters based on the gradient of some performance measure $J(\bm{\theta})$ with respect to the policy parameters $\bm{\theta}$. Seeking to maximize the performance, their updates approximate the gradient ascent of $J$
\begin{equation}
	\bm{\theta_{k+1}}=\bm{\theta_k}+\alpha \widehat{\nabla J(\bm{\theta_k})},
\end{equation}
where $\widehat{\nabla J(\bm{\theta_k})}$ is a stochastic estimation, the expectation of which approximates the gradient of $J(\bm{\theta_k})$ with respect to $\bm{\theta_k}$.

We start from the undiscounted case where $\gamma =1$. Let $\pi_{\Omega,\theta}$ denote the meta-policy parameterized by $\theta$ and $\beta_{\omega, \vartheta}$, the termination function of $\omega$ parameterized by $\vartheta$. In our case, the intra-option policies $\pi_\omega$ are deterministic and do not need to be learned from scratch. We define the value of the start state of each episode as the performance measure 
\begin{equation}
	J(\theta) \coloneqq V_\Omega(s_0),
\end{equation}
where $V_\Omega(s)$ is the value function under the meta-policy $\pi_\Omega$. To keep the notation simple, we leave it implicit that $\pi_\Omega$ is parameterized by $\theta$ and the gradients are also with respect to $\theta$. The gradient of the state-value function can be rewritten in terms of action-value function as
\begin{equation} \label{eq:pg1}
	\begin{split}
		\nabla V_\Omega(s) &= \nabla[\sum_{\omega} \pi_\Omega(\omega|s)Q_\Omega(s,\omega)] \\
						   &= \sum_{\omega}[\nabla \pi_\Omega(\omega|s)Q_\Omega(s,\omega) + \pi_\Omega(\omega|s) \nabla Q_\Omega(s,\omega)].  					   
	\end{split}
\end{equation}

Since $Q_\Omega(s,\omega) = R(s,\omega) + \sum_{s'}P(s'|s,\omega) V_\Omega(s')$, where $P(s'|s,\omega)=\sum^{\infty}_{\tau = 1}P(s', \tau|s,w)$, Eq. (\ref{eq:pg1}) can be written in
\begin{equation} \label{eq:pg2}
	\begin{split}
		\nabla V_\Omega(s) &= \sum_{\omega}[\nabla \pi_\Omega(\omega|s)Q_\Omega(s,\omega) \\
						   & \qquad + \pi_\Omega(\omega|s)  \sum_{s'} P(s'|s,\omega) \nabla V_\Omega(s')] \\
						   &= \sum_{\omega}[\nabla \pi_\Omega(\omega|s)Q_\Omega(s,\omega) + \pi_\Omega(\omega|s) \sum_{s'} P(s'|s,\omega) \\
						   & \qquad \sum_{\omega'}[\nabla \pi_\Omega(\omega'|s')Q_\Omega(s',\omega') \\
						   & \qquad + \pi_\Omega(\omega'|s') \sum_{s''} P(s''|s',\omega') \nabla V_\Omega (s'')] ] \\
						   &= \sum_{x\in S}\sum^{\infty}_{k=0}Pr(s\to x, k, \Omega) \sum_{\omega} \nabla\pi_\Omega(\omega|x)Q_\Omega(x,\omega),						   	
	\end{split}
\end{equation}
after repeated unrolling \cite{sutton1998reinforcement}, where $Pr(s\to x, k, \Omega) $ represents the probability of transitioning from state $s$ to state $x$ in $k$ option steps under policy $\pi_\Omega$. Thus
\begin{equation} \label{eq:pg3}
	\begin{split}
		\nabla J(\theta) &= \nabla V_\Omega(s_0)\\
						 &= \sum_{s}\left( \sum^{\infty}_{k=0}Pr(s_0\to s, k, \Omega)\right)  \sum_{\omega} \nabla\pi_\Omega(\omega|s)Q_\Omega(s,\omega)	\\	
						 &\propto \sum_{s} \mu(s) \sum_{\omega} \nabla\pi_\Omega(\omega|s)Q_\Omega(s,\omega) \\
						 &= \mathbb{E}_\pi\left[\sum_{\omega} \nabla\pi_\Omega(\omega|s_t)Q_\Omega(s_t,\omega)\right] \\
						 &= \mathbb{E}_\pi\left[\sum_{\omega} \pi_\Omega(\omega|s_t)  \frac{\nabla\pi_\Omega(\omega|s_t)}{\pi_\Omega(\omega|s_t)}Q_\Omega(s_t,\omega)\right] \\
						 &= \mathbb{E}_\pi\left[\frac{\nabla\pi_\Omega(\omega_t|s_t)}{\pi_\Omega(\omega_t|s_t)}Q_\Omega(s_t,\omega_t)\right] \\
						 &= \mathbb{E}_\pi\left[\nabla \text{ln}\pi_\Omega(\omega_t|s_t)Q_\Omega(s_t,\omega_t)\right], \\
	\end{split}
\end{equation}
where $\mu(s)$ here is the on-policy state distribution under $\pi_\Omega$, $s_t$ and $\omega_t$ are sample state and option. Note that both $s_t$ and $\omega_t$ lie on the option scale, i.e. they are sampled at the time step when the option is initialized. 
In this way, the expectation of the sample gradient is proportional to the actual gradient of the performance measure (which is the value function in our case) with respect to the parameter and thus the meta-policy can be updated via $\theta_{k+1}=\theta_k+\alpha \widehat{\nabla J(\theta_k)}=\theta_k+\alpha \nabla_\theta \text{ln} \pi_{\Omega,\theta}(\omega_t|s_t) Q_\Omega(s_t,\omega_t)$. More generally, we use a critic $V_{\theta_v}$ to approximate the state-value function, parameterized by $\theta_v$, to reduce the variance and speed up the learning process.  
We generalize the previous undiscounted version to discounted case ($\gamma\neq 1$). As Thomas showed in \cite{thomas2014bias}, the discount factor makes the usual policy gradient estimator biased. However, correcting for this discrepancy also incurs data inefficiency. As discussed in \cite{bacon2017option}, we build our model based on the policy gradient estimator for simplicity.
Suppose the  duration of the option $\omega_t$ is $\tau$, i.e. $\omega_t$ lasts for $\tau$ time steps, then $R_{t:t+\tau}$ is the discounted accumulated return during $t$ to $t+\tau$. The  parameters of the meta-policy $\pi_{\Omega,\theta}$ is updated at the option scale:
\begin{equation}
	\begin{split}
	\theta_{k+1}= &\theta_k+ \alpha_\theta (R_{t:t+\tau} +\gamma^\tau V_\Omega(s_{t+\tau}) \\
						 & - V_\Omega(s_t)) \nabla_{\theta} \text{ln}\pi_{\Omega,\theta}(\omega_t|s_t). \\
	\end{split}
\end{equation}

\begin{figure}[!t]
	\removelatexerror
	\begin{algorithm}[H]
		\caption{Option-Interruption algorithm with policy gradient update} \label{algOI}
		Initialize global step counter $T \gets 1 $\;
		Initialize $\pi_{\Omega}$ weight $\theta$, $\beta_\omega$ weight $\vartheta$, $V_\Omega$ weight $\theta_v$\;
		\Repeat{$T> T_{max}$}
		{
			Initialize episode step counter $t \gets 1 $\;
			Get state $s_t$\;
			\Repeat{\normalfont{terminal} $s_t$ or $t>t_{max}$}
			{
				Reset gradients $d\theta \gets 0$, $d\theta_v \gets 0$, $d\vartheta \gets 0$\;
				$t_{start} \gets t$\;
				Choose $\omega_t$ according to $\pi_\Omega(\omega_t|s_t)$\;
				\Repeat{$\beta(s_t)$ \normalfont{terminates or terminal} $s_t$}
				{
					Choose $a_t$ according to $\pi_\omega(a_t|s_t)$\;
					Receive reward $r_t$ and new state $s_{t+1}$\;
					$t \gets t+1$, $T \gets T+1$\;
				}
				$
				R = \begin{cases}
				0 & \text{for terminal $s_t$,} \\
				V(s_t) & \text{for non-terminal $s_t$.} \\
				\end{cases}
				$\\
				\For{$i \in \{t-1,...t_{start}\}$}
				{
					$R = r_i +\gamma R$\;
					$d\vartheta \gets d\vartheta + \frac{\partial\beta_\omega(s_{i})}{\partial \vartheta} (R - V(s_{i}))$\;
					$d\theta_v \gets d\theta_v + \frac{\partial(R - V(s_i))^2}{\partial \theta_v}$\;
					
				}
				$d\theta \gets d\theta + \nabla_\theta \text{ln}\pi_\Omega(\omega_{t_{start}}|s_{t_{start}}) (R - V(s_{t_{start}}))$;\
				Update $\theta$, $\theta_v$, $\vartheta$ using $d\theta$, $d\theta_v$, $d\vartheta$
			}
		}
	\end{algorithm}
\end{figure}

Interrupting options before they would finish naturally according to their termination conditions endows the agent with the flexibility to switch options when necessary. As derived in \cite{bacon2017option} the parameters of the termination function $\beta_{\omega,\vartheta}$ can be updated at each action time step as follows:
\begin{equation}
	\vartheta_{k+1}=\vartheta_k - \alpha_\vartheta \left(R_{t} + \gamma V_\Omega(s_{t+1}) - V_\Omega(s_{t})\right) \nabla_\vartheta \beta_{\omega,\vartheta}(s_{t}).
\end{equation}

The whole Option-Interruption algorithm with policy gradient update is presented in Algorithm \ref{algOI}. Notice that the meta-policy $\pi_\Omega$, the approximated state-value function $V_\Omega$ and the option termination funciton $\beta_\omega$ are updated at different temporal scales.

\section{EXPERIMENTS} \label{sec:exp}
	\subsection{Four-room Navigation}
\begin{figure} 
		\centering 
		\subfloat[]{\includegraphics[height=120pt]{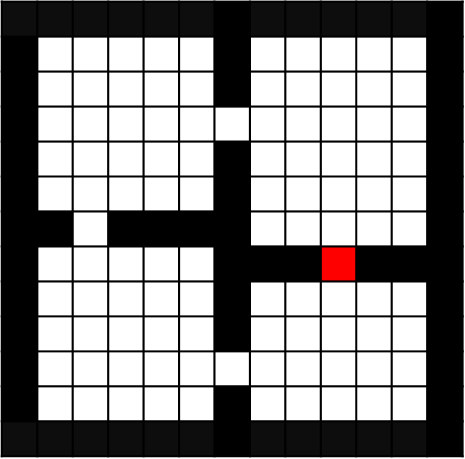}}
		\qquad
		\subfloat[]{\includegraphics[height=120pt]{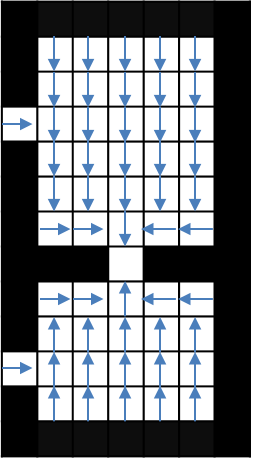}}
		\caption{(a) Four-room grid world environment. The goal is marked in red. (b) The initialization set and the intra-option policy of east hallway option.} \label{fig:fourrooms}
	\end{figure}

	To verify the effectiveness of our algorithm, firstly we consider the navigation problem in a grid world environment of four rooms as shown in \cite{sutton1999between}. The four-room environment is depicted in Fig. \ref{fig:fourrooms}(a), where the cells of the grid correspond to the states of the environment. There are four hallways connecting adjacent rooms and our goal is the east hallway, marked in red in the figure. At the beginning of each episode, the agent is placed at a random location. From any state, it can perform one of four primitive actions： \emph{up}, \emph{down}, \emph{left} or \emph{right}. In addition, primitive movements can fail with a probability of $1/3$, in which case the agent randomly transit to one of the empty adjacent cells. The wall, represented by black cell, is not accessible. If an agent hits the wall, then it will remain in the same cell.

	In our environment, there are four \emph{options} corresponding to four hallways. At any cell, only the two options that lead the agent to the hallways of the current room are available. Fig. \ref{fig:fourrooms}(b) gives an example, showing the initialization set $I_\omega$ of the east hallway option along with its policy $\pi_\omega$, following the shortest path within the room to its target hallway. Besides, the reward is always $0$ on all state transitions except transiting to the goal. 
	
		\begin{table}[!th]
		\setlength{\tabcolsep}{1.0em}
		\caption{Comparison result of Four-room Navigation. The columns $t=i$ refer to the episode length after trained $i$  times. The columns $n<i$ represent the time when the episode length is under $i$. }
		\label{table:fourroom}
		\centering
		\begin{tabular}{c|c|c|c|c}
			\hline
			&$t=0$&$t=1000$&$n<50$ & $n<20$  \\
			\hline
			AC &362.22&11.28&65&226 \\
			OC &357.69&15.45&119&622 \\
			Our method &62.59&9.97&3&21 \\
			No Interruption &22.56&10.85&-&4\\
			\hline
			
		\end{tabular}
	\end{table}
	
	We compare our method with four-option Option-Critic (OC) architecture \cite{bacon2017option} which learns both policy over options $\pi_\Omega$ and option policies $\pi_\omega$ from scratch. We also implement Actor-Critic (AC) method at the primitive action level. In all methods, $\pi_\Omega$ is parameterized with Boltzmann distribution and the termination function $\beta_\omega$ is parameterized with sigmoid functions. The discount factor $\gamma$ is $0.99$, and all value functions are updated by one-step learning. All the weights are initialized to zero. 
	
	\begin{figure}
		\centering
		\includegraphics[height=120pt]{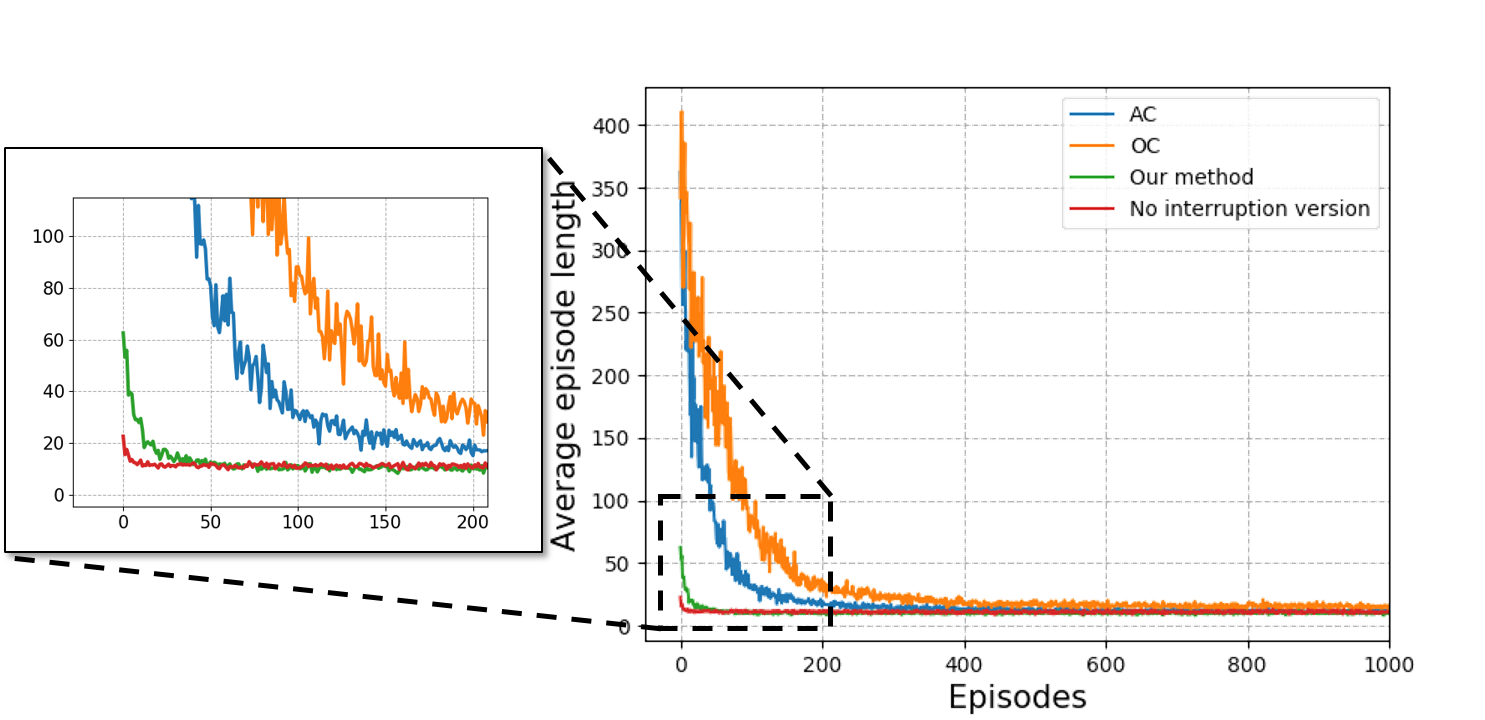}
		\caption{Average episode length. All curves are the averaged result over $100$ runs.} \label{fig:four-exp1}
	\end{figure}
	
	As shown in Fig. \ref{fig:four-exp1}, during the training process the average episode length of our method converges much faster than that of OC and AC, which learn primitive actions from scratch, indicating that the involvement of prior knowledge can indeed help to speed up the learning process. Another observation is that at the early stage of the training process, the average episode length of our method is remarkably lower as displayed in TABLE \ref{table:fourroom}, which suggests that the options involving temporal abstraction are the key to reduce the searching space. Such advantage is extremely important to real robots since it will constrain the robot's behavior and accordingly protect it from unexpected damage when deployed in the real world. 
	
\begin{figure}
	 	\centering
		\subfloat[]{\includegraphics[height=130pt]{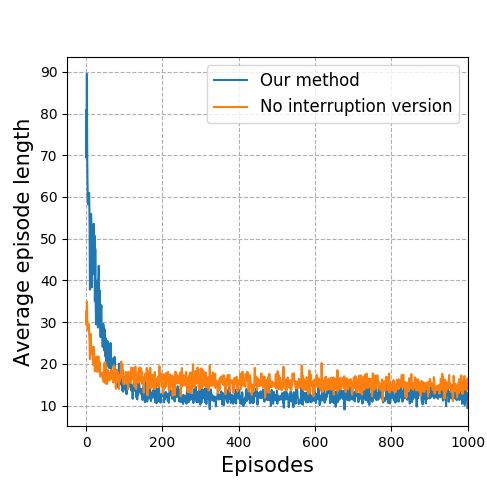}}
		\subfloat[]{\includegraphics[height=130pt]{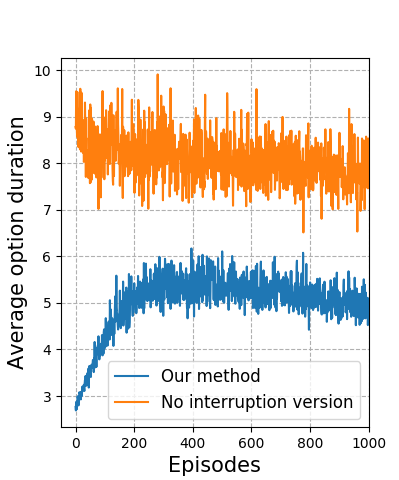}}
	 	\caption{Experiment results in the dynamically blocked environment. (a) Average episode length. Our method with interruption has a lower length. (b) Average option duration. All curves are averaged over $100$ runs.} \label{fig:four-exp2}
	 \end{figure}

	It is noted that the effect of termination function in this typical setting is not fully validated since the optimal policy actually does not contain any termination. Hence we design a variant version of Four-room navigation problem, where one of the three hallways (hallways except the goal) will be randomly blocked for $\tau$ time steps where $\tau$ is a random variable uniformly distributed in $(0, 20]$. We compare our method with the none-interruption version, i.e. the option will be terminated only when reaching the target hallway, and the result at the learning stage is shown in Fig. \ref{fig:four-exp2}. After around $200$ episodes the episode length of the version without termination function is higher than the version with termination function and the average option duration is much shorter when equipped with the termination function, from which we can easily recognize the effect of the interruption mechanism. The interruption mechanism endows the agent with the flexibility to switch the policy timely, and in our case, the agent will interrupt the ongoing option if the target hallway is blocked. 
	
	\subsection{Autonomous Exploration in Indoor Environments} \label{sec:exporation}
	Autonomously exploring an unknown environment is an essential task for mobile robots, where the agent is expected to find out a safe path to cover the whole map with the constraint of reducing the path cost as much as possible \cite{Zhu2018Deep}. In this subsection, we perform the exploration task in an indoor environment and compare our Option-Interruption architecture with the typical DRL.
	
	In terms of learning based exploration methods, the agent is expected to summarize the strategy based on its experience, since the indoor layouts of houses are well structured and contain rich spatial information. At each episode, the mobile range-sensing robot starts from a random location and explores the environment in a discrete manner. Every time step it moves a fixed length and the episode ends when the whole area is covered. The key challenges for this task are: 1) the agent must learn to avoid obstacles; 2) the agent is expected to learn to move towards the unknown areas.
    
    \begin{figure}
    	\centering
    	\includegraphics[width=200pt]{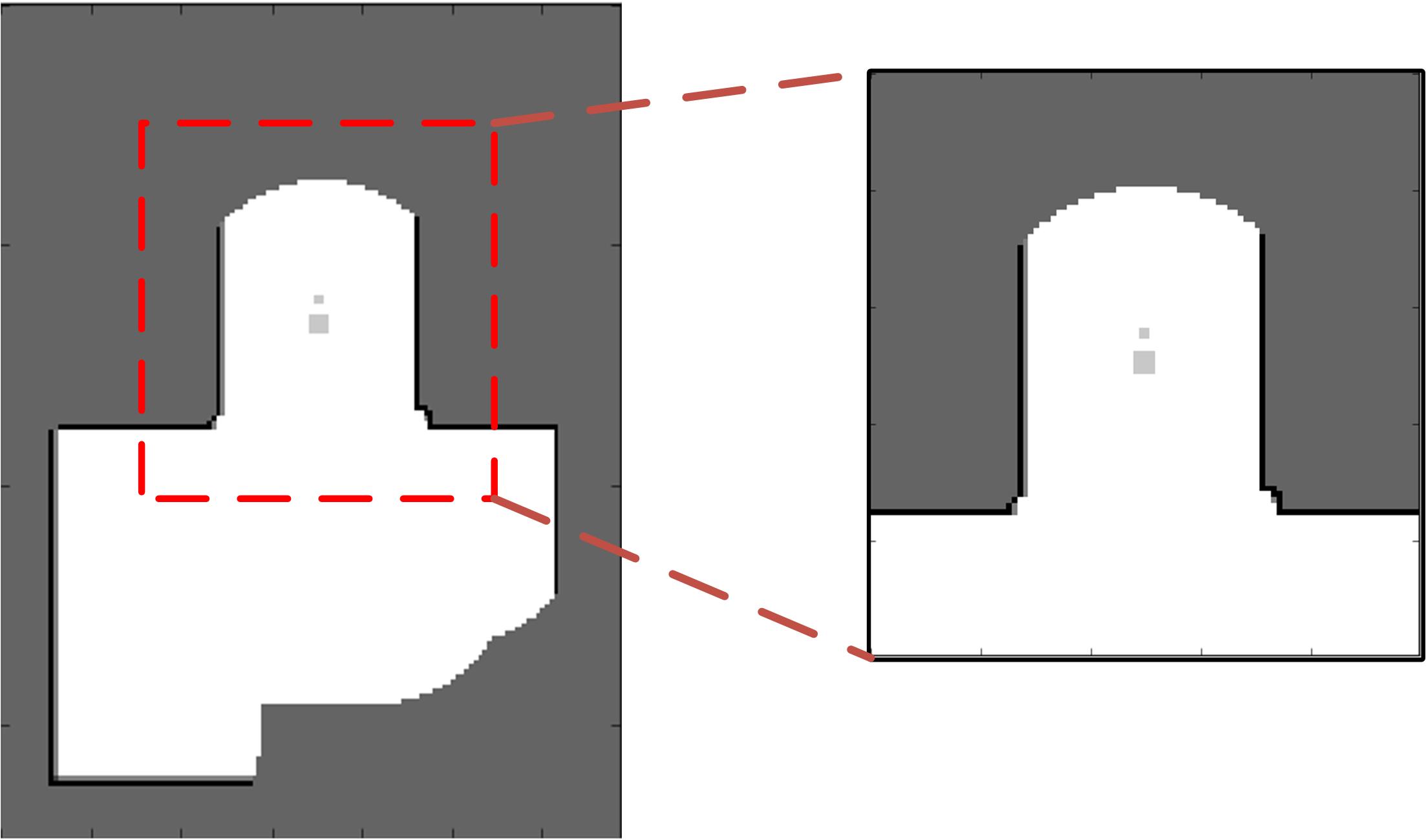}
    	\caption{The input of the network. White pixels mean free space, gray pixels are unknown areas and black pixels represent obstacles. The input is a $80 \times 80$  pixels image patch centered at the current location of the robot.} \label{fig:submap}
    \end{figure}
    
    In our experiment, the state $s_t$ is a $80 \times 80$ pixels image patch centered at the current location of the robot, as shown in Fig. \ref{fig:submap}. The robot is drawn at the center of $s_t$, to indicate its orientation. The goal of robot exploration task is to 1) cover the house in a low path length; 2) guarantee obstacle avoidance. In term of the reward function, ideally three signals are enough to reflect the aforementioned goals: 1) a time penalty at each step to urge the agent to finish	the task; 2) a success reward when completing the task; 3) a collision penalty when hitting walls. However, the reward function $R$ defined in this way is too difficult for the agent to learn, due to the sparsity of the positive signals. Hence we define $R$ in an informative way: the newly explored area is taken into consideration to encourage the agent to collect more information at each step. Our reward function at time step t is defined as
	\begin{equation}    
	r_t = \begin{cases}
	s_{t} c_{s} + p_{time} &      \text{if no collision happens at $t$,} \\
	p_{collision} & \text{if collision happens at $t$,} \\
	r_{success} &   \text{if agent have explored all area.}
	\end{cases}
	\end{equation}
	where $s_{t}$ describes newly covered area (presented by the number of pixels). $c_{s}$ is a constant coefficient for scaling. $p_{time}$ and $p_{collision}$ represent the time penalty and the collision penalty. And $r_{success}$ is the reward if the agent completes the exploration task. 
	
	The actions of typical RL are four directions, \emph{up, down, left} and \emph{right} with fixed step length. As for our hierarchical structure, the option is specified as one-step action same as typical RL but equipped with the obstacle avoidance ability, i.e. for each option, the initialization set $I_\omega$ only contains the available  adjacent free cells. In this way, the agent would not need to learn to avoid walls.
	
\begin{figure}
		\centering
		\includegraphics[width=200pt]{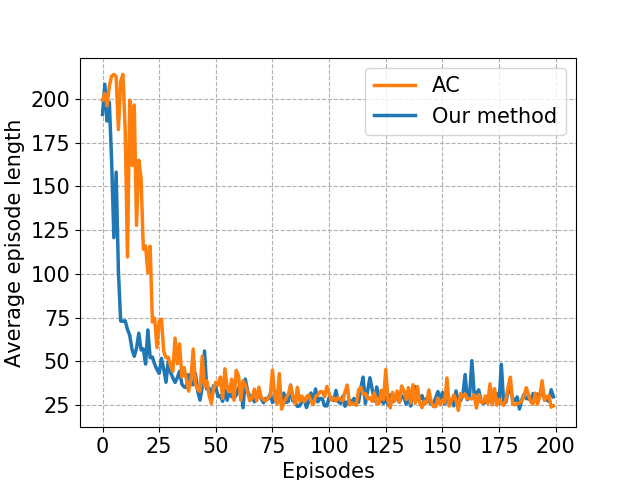}
		\caption{Average episode length at training stage. All curves are averaged over $16$ parallel threads.} \label{fig:exploration-curve}
	\end{figure}

	For the purpose of enabling the robot exploring environments autonomously and efficiently, we build the Asynchronous Advantage Actor-Critic (A3C) network \cite{mnih2016asynchronous} with $16$ parallel workers and the network details are as follows:
	\begin{enumerate}
		\item A convolution layer with $(k8, c32, s4)$;		
		\item A convolution layer with $(k4, c64, s2)$;		
		\item A convolution layer with $(k3, c64, s1)$;
		\item A fully connected layer with $512$ units;
		\item A policy head $\pi$ and a value head $v$ for RL, \\
			  or a meta-policy head $\pi_\Omega$, a value head $v_\Omega$ and $n$ termination function $\beta_{\omega}$ for HRL;
	\end{enumerate}
	where $(k8, c32, s4)$ refers to kernel size $8\times 8$, number of outputs $32$, stride $4$. With an emphasis on the influence on training process, we trained our agent on one map without testing its generalization ability. And the result is shown in Fig. \ref{fig:exploration-curve}.
	
	As we can see, our method converges much faster than Actor-Critic (AC). The largest distance between the two curves lies in $14$th episode which is $143$. This clearly proves that by giving human knowledge the training process can be significantly sped up. And our result can be further improved by implementing other existing methods.

\section{CONCLUSION}
In summary, this paper proposes an Option-Interruption architecture that embeds existing methods into a hierarchical reinforcement learning structure. On the basis of the existing methods, the search space is considerably reduced and hence the training process is significantly sped up. On the other hand, the interruption mechanism provides the flexibility to the changing of the external world which existing methods do not hold. The experiment shows the efficiency of our architecture given proper human knowledge. 

At the same time, there is still some future work to do. For example, the training process of the termination function can be further investigated. As displayed in Fig. \ref{fig:four-exp2}(a), although the final performance of no interruption version is worse, it performs well at the initial training stage, which means it may be a choice to disable the termination function at first to protect our fragile robot when we are training a real robot. Another consideration is to apply our method to real robots, which leaves us a lot of work to do.

\bibliographystyle{IEEEtran}
\bibliography{robio2018}

\end{document}